\documentclass[11pt]{article}

\usepackage[final]{acl}

\usepackage{fontspec}
\usepackage{polyglossia}
\usepackage{times}
\usepackage{latexsym}

\usepackage[T1]{fontenc}
\newfontfamily\devanagarifont{NotoSansDevanagari.ttf}[Script=Devanagari]
\newcommand{\devanagari}[1]{{\devanagarifont #1}}
\usepackage{array}
\usepackage{enumitem}
\usepackage{todonotes}

\usepackage{pifont}
\definecolor{mediumgreen}{HTML}{13B015}

\definecolor{g}{RGB}{0,0,204}     
\definecolor{l}{RGB}{0,102,0}    
\definecolor{h}{RGB}{153,0,153} 
\definecolor{de}{RGB}{204,0,0}  

\usepackage{multicol}
\usepackage{makecell}

\colorlet{an1}{blue!50}
\colorlet{an2}{orange}
\colorlet{an3}{yellow!80!orange}
\colorlet{anbox1}{yellow!50!orange}
\colorlet{anbox2}{cyan}

\usepackage{microtype}
\usepackage{float}

\usepackage{inconsolata}

\usepackage{graphicx}
\usepackage{enumitem}
\usepackage{booktabs}
\usepackage{longtable}
\usepackage{amssymb}
\usepackage{amsmath}
\usepackage{array}
\usepackage[linesnumbered,ruled,vlined]{algorithm2e}

\usepackage{multirow}
\usepackage{tcolorbox} 
\tcbuselibrary{breakable}
\usepackage{caption} 

\title{Padamitra: Grounded Glossary Generation for Classical Sanskrit}

\author{Manoj Balaji Jagadeeshan \And Sai Pragnaan Marala \And Pawan Goyal\\
\AND
Indian Institute of Technology Kharagpur}

\begin{document}
\maketitle
\begin{abstract}
    We introduce grounded glossary generation, a structured task requiring models to recover semantically meaningful Sanskrit phrases and produce translation-grounded meanings from a \'{s}loka - translation pair, formalizing the traditional p\={a}\d{t}ha commentary practice as an evaluable NLP objective. We construct a benchmark of 31,316 \'{s}loka - translation – glossary triples from the V\={a}lm\={i}ki R\={a}m\={a}ya\d{n}a and \'{S}r\={i}mad Bh\={a}gavatam, paired with two metrics: Jaccard for phrase recovery and Meaning Faithfulness for semantic consistency. Across zero-shot, few-shot, and instruction fine-tuned variants of Gemma-3n-E4B, Gemma-3-12B, Phi-4, and Qwen3.5-9B, instruction fine-tuning substantially outperforms prompting, while explicit segmentation yields gains. Error analysis identifies over-segmentation of sandhi and sam\={a}sa compounds as the dominant failure mode, pointing to morphological modeling as the key bottleneck for faithful Sanskrit lexical decomposition. Data\footnote{\url{https://huggingface.co/collections/sanganaka/padamitra-glossary-generation}} and Code\footnote{\url{https://github.com/sanganaka-iitkgp/Padamitra-Glossary-Generation}} are available.
\end{abstract}
\section{Introduction}
\label{sec:intro}
 
Classical Sanskrit literature poses significant challenges for interpretation due to its morphological richness and flexible syntax. In particular, phonological fusion across word boundaries (\textit{sandhi}) and systematic nominal compounding (\textit{sam\={a}sa}) often produce surface forms whose segmentation into semantically meaningful units is itself a non-trivial task \cite{huet2009,hellwig-nehrdich-2018-sanskrit}.
Classical commentators addressed this through \textit{p\={a}\d{t}ha} commentary - a glossary-like annotation that maps each resolved phrase in a verse (\textit{\'{s}loka}) to its contextual meaning - a tradition that underlies modern pedagogical resources such as the IITK Valmiki R\={a}m\={a}ya\d{n}a\footnote{\url{https://valmiki.iitk.ac.in}} and Vedabase\footnote{\url{https://vedabase.io/en/library/sb/}}.
We formalize this tradition as \textbf{grounded glossary generation}: given a Sanskrit \textit{\'{s}loka} $s$ and its authoritative English translation $t$, the task is to produce a structured glossary $\mathcal{G} = \{(k_i, v_i)\}_{i=1}^{m}$, where each key $k_i$ is a semantically coherent Sanskrit phrase and each value $v_i$ is a meaning grounded in $t$.
\begin{figure}[t]
    \centering
    \includegraphics[width=0.48\textwidth]{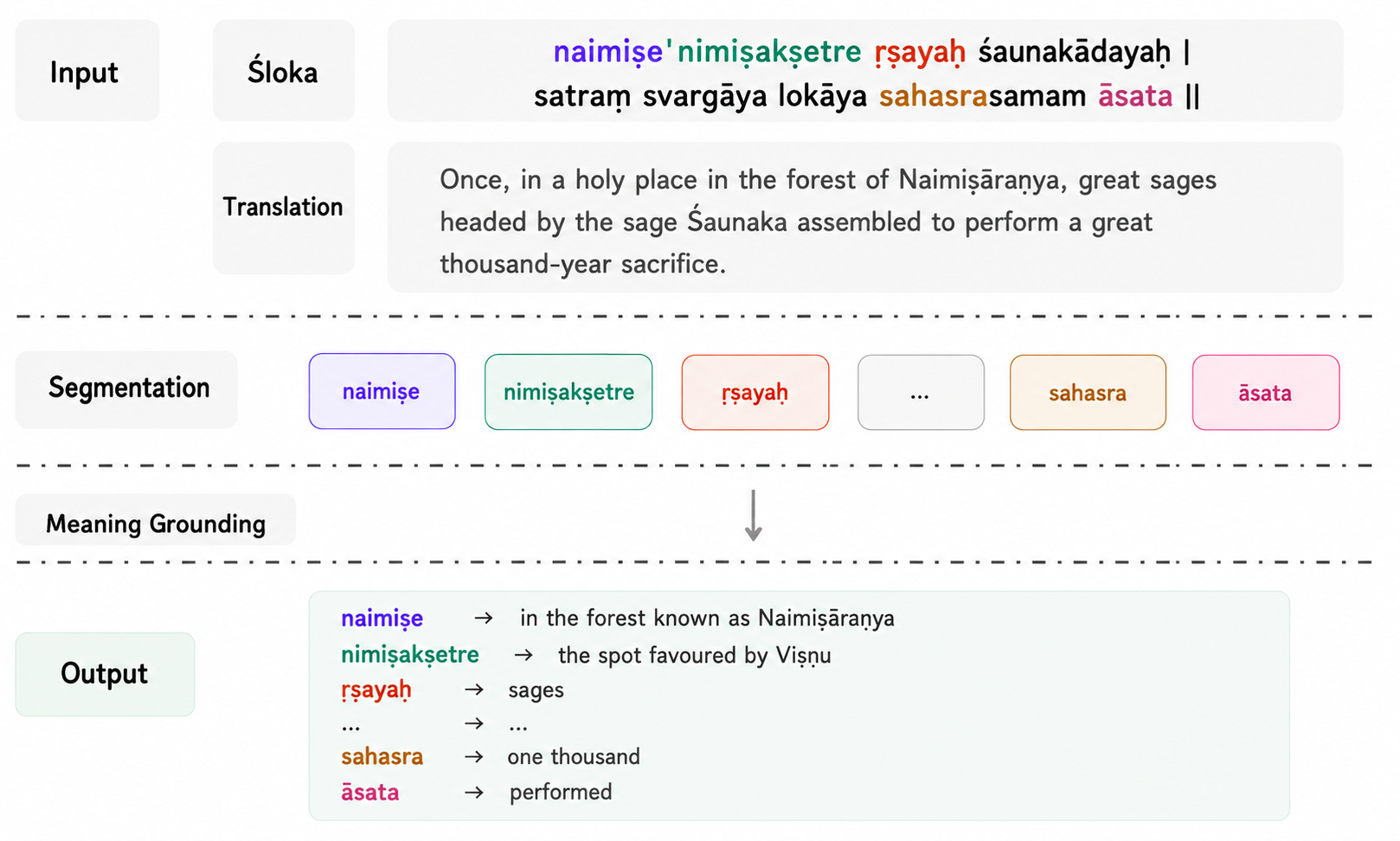}
    \caption{Grounded glossary generation. From a \'{s}loka–translation pair, the model recovers sandhi- and sam\={a}sa-resolved Sanskrit phrases and assigns each a meaning grounded in the provided translation, yielding a structured glossary. }
    \label{fig:task}
\end{figure}
Crucially, grounding in the translation distinguishes this task from unconstrained lexicographic gloss generation: the meanings must be \emph{faithful} to the provided translation, not merely plausible in isolation.
This formulation makes interpretable lexical decomposition an explicit, evaluable objective rather than an opaque by-product of end-to-end translation.

Prior work in Sanskrit NLP has largely targeted individual sub-problems: sandhi splitting, morphological analysis, and segmentation~\cite{huet2009,kulkarni2010,krishna2017,hellwig-nehrdich-2018-sanskrit}; cross-lingual retrieval~\cite{jagadeeshan-etal-2025-anveshana}; entity resolution~\cite{sarkar-etal-2025-mahanama}; \textit{anvaya}~\cite{krishna-etal-2019-poetry,das-etal-2025-still}; and poetry generation~\cite{jagadeeshan-etal-2026-chandomitra,jagadeeshan2026pingalaprosodyawaredecodingsanskrit}; \citet{krishna-etal-2020-graph} address several of these jointly in a single graph-based framework. Related work has also explored Sanskrit dependency parsing~\cite{kulkarni2010}, Sanskrit--English neural machine translation~\cite{aralikatte2021, nehrdich2026mitrasamgrahacomprehensiveclassicalsanskrit}, multilingual LLMs for low-resource languages~\cite{ahuja2023}, and bilingual word alignment using FastAlign~\cite{dyer-etal-2013-simple}. However, no prior work directly addresses grounded phrase-level glossary generation as a unified task combining segmentation, semantic grounding, and lexicographic faithfulness.\\
Four gaps motivate this work. First, token-level aligners such as FastAlign cannot model Sanskrit's fusional morphology, where one surface token may span several glossary keys. Second, instruction-tuned LLMs handle structured prediction well via in-context learning~\cite{brown2020,wei2022}, but their behavior on a task coupling sandhi resolution, boundary detection, and translation-grounded meaning assignment is untested. Third, MT metrics such as BLEU \cite{papineni-etal-2002-bleu} and chrF \cite{popovic-2015-chrf} measure sentence-level fluency, not phrase-level key recovery or per-entry semantic faithfulness. Fourth, whether explicit segmentation supervision helps over end-to-end learning is unestablished.\\
We address these gaps with a benchmark of 31{,}316 \textit{\'{s}loka}--translation--glossary triples from the V\={a}lm\={\i}ki R\={a}m\={a}ya\d{n}a and \'{S}r\={\i}mad Bh\={a}gavatam, and two metrics: \textbf{Jaccard} for glossary key recovery and \textbf{Meaning Faithfulness} for translation-grounded semantic consistency. Across zero-shot, few-shot, and instruction fine-tuned variants of \texttt{gemma-3n-E4B-it}, \texttt{gemma-3-12b-it}, \texttt{phi-4}~(14B), and \texttt{Qwen3.5-9B}, instruction fine-tuning (IFT) substantially outperforms prompting. Ablations show that predicted segmentation adds little, though gold segmentation raises Meaning Faithfulness from 0.787 to 0.872, and that IFT models degrade under word-order perturbation. Error analysis identifies over-segmentation of \textit{sandhi} and \textit{sam\={a}sa} compounds as the dominant failure mode.

\section{Problem Statement}

Given a Sanskrit \'{s}loka(s) and its English translation (t), the goal is to generate a grounded glossary \(
G = {(k_i, v_i)}_{i=1}^{m}    
\) where each \(k_i\) is a semantically meaningful Sanskrit phrase and \(v_i\) is its meaning grounded in the translation (t).\\
Unlike machine translation \cite{koehn2003statistical}, which targets fluent output over automatically induced phrase pairs, bilingual lexicon induction \cite{conneau2018word, artetxe2017learning}, which derives word-level equivalents from corpus-level statistics, or definition modeling \cite{noraset2017definition}, which generates context-free glosses, grounded glossary generation jointly requires recovering morphologically complex phrase boundaries and producing meanings faithful to a specific provided translation. The generated glossary must remain grounded in the provided translation, capturing interpretable lexical units as structured phrase-meaning pairs while avoiding unsupported or hallucinated interpretations.
\begin{tcolorbox}[title=Example Illustration,colback=white, colframe=brown]
\small
\textbf{Sanskrit \'{s}loka:} \devanagari{नैमिषेऽनिमिषक्षेत्रे ऋषय: शौनकादय: ।} \\
\devanagari{ सत्रं स्वर्गायलोकाय सहस्रसममासत ॥} \\
\textbf{English Translation:} \\
Once, in a holy place in the forest of Naimi\d{s}\={a}ra\d{n}ya, great sages headed by the sage \'{S}aunaka assembled to perform a great thousand-year sacrifice for the satisfaction of the Lord and His devotees. \\
\textbf{Glossary:} \\
\devanagari{नैमिषे} $\rightarrow$ in the forest known as Naimi\d{s}\={a}ra\d{n}ya \devanagari{अनिमिष-क्षेत्रे} $\rightarrow$ the spot which is especially a favorite of Vi\d{s}\d{n}u (who does not close His eyelids) \devanagari{ऋषयः} $\rightarrow$ sages \devanagari{शौनक-आदयः} $\rightarrow$ headed by the sage \'{S}aunaka \devanagari{सत्रम्} $\rightarrow$ sacrifice \devanagari{स्वर्गाय} $\rightarrow$ the Lord who is glorified in heaven \devanagari{लोकाय} $\rightarrow$ and for the devotees who are always in touch with the Lord \devanagari{सहस्र} $\rightarrow$ one thousand, \devanagari{समम्} $\rightarrow$ years \devanagari{आसत} $\rightarrow$ performed.
\end{tcolorbox}

\section{Dataset}

\subsection{Data Sources}
The dataset is curated from two publicly available educational resources: R\={a}m\={a}ya\d{n}a and \'{S}r\={\i}mad Bh\={a}gavatam, and use it for research purposes. Each instance contains:
\begin{center}
\small
    \textit{(Sanskrit \'{s}loka, English translation, a glossary mapping Sanskrit phrases) }
\end{center}

\subsection{Data Preparation}
We parse semi-structured glossary annotations into structured key-value dictionaries and apply light normalization, including Unicode and whitespace normalization, while preserving the original linguistic structure of the text. \\
The dataset is divided into train, validation, and test splits:
\textbf{Train:} 25,050, \textbf{Validation:} 3,133, \textbf{Test:} 3,133.

To reduce leakage, contiguous verses and near-duplicate glossary entries are prevented from appearing across splits. Overlaping gloss key between Train, Validation and Test is shown in \autoref{tab:overlap-gloss-key}.

\section{Evaluation Metrics}
\subsection{Jaccard}

To measure glossary key recovery, we compute an overlap-based Jaccard score between the set of generated glossary keys $(G_p)$ and reference glossary keys $(G_r)$:
\[
    \text{Jaccard} = \frac{|G_p \cap G_r|}{|G_p \cup G_r|}
\]
where each glossary key corresponds to a Sanskrit phrase segment. This metric evaluates how accurately the model recovers the reference semantic units. 
\subsection{Meaning Faithfulness}

Meaning Faithfulness measures semantic consistency between generated and
reference glossary entries while penalizing segmentation errors in the keys.

For each generated key $k_p$, we find the closest reference key $k_r$ by
normalized Levenshtein similarity, computed with
RapidFuzz\footnote{\url{https://github.com/rapidfuzz/RapidFuzz}}~\cite{max_bachmann_2025_15133267}.
We write this similarity as $\text{KeySim}(k_p,k_r) \in [0,1]$ and treat the pair as matchable only above a threshold of $0.7$; below it, $\text{KeySim} = 0$. For matched pairs, we compare the corresponding meanings using cosine similarity between BAAI/bge-base-en-v1.5~\cite{bge_embedding} embeddings, obtained via SentenceTransformers~\cite{reimers-2019-sentence-bert,reimers-2020-multilingual-sentence-bert}:
\[
  \text{ValSim}(v_p,v_r) = \cos\!\big(\phi(v_p), \phi(v_r)\big),
\]
where $\phi(\cdot)$ is the embedding function. Each generated entry contributes the product
{
\small
\[
  \text{Contribution} = \text{KeySim}(k_p,k_r) \cdot \text{ValSim}(v_p,v_r),
\]
}
and the score for a \'{s}loka is the mean contribution over all generated entries; the reported score is the mean over the evaluation set. Averaging over \emph{generated} entries penalizes spurious keys directly: an unmatched key contributes zero and drags the mean down. \autoref{appendix:metric-intuition} provides intuition behind the design of this metric.

\section{Methodology}
\subsection{Models Used}
For the task of Glossary generation, we consider the models: \texttt{gemma-3n-E4B-it} \cite{gemma_3n_2025}, \texttt{gemma-3-12b-it} \cite{gemma_2025}, \texttt{phi-4} \cite{abdin2024phi4technicalreport}, \texttt{Qwen3.5-9B} \cite{qwen3.5}. The models translation capability is discussed in \autoref{appendix:model-translation-capability}.

\subsection{Experimental Setup}
\paragraph{Baselines.} As a non-LLM baseline we use \texttt{FastAlign}~\cite{dyer-etal-2013-simple}, a lightweight statistical word aligner based on IBM Model~2~\cite{brown-etal-1993-mathematics} that learns token-level correspondences from parallel corpora; we extract the glossary from the alignments induced between each \'{s}loka and its English translation.\\
We additionally evaluate ByT5-Sanskrit \cite{nehrdich-etal-2024-one} in two configurations: (i) a single seq2seq model fine-tuned end-to-end to generate the glossary directly from the \'{s}loka and translation, and (ii) a two-stage pipeline in which one model is fine-tuned for segmentation (\'{s}loka $\to$ gloss keys) and a second generates the glossary from the predicted segments and the reference translation.

\noindent \textbf{In-Context Learning:} We experiment whether the a language model learn to generate the glossary with increased number of examples via zero-shot, few-shot learning with 1, 5 and 10 examples. These examples are available in \autoref{appendix:n-shot-example}. The system and user prompt used for N-shot generation is available in \autoref{box:system_prompt}. 

\noindent \textbf{Instruction Fine-tuning (IFT):} We instruction fine-tune language models using the aforementioned user prompt.

Settings for IFT is documented in \autoref{appendix:llm_hyperparameter}.

\section{Result}
\subsection{Quantitative Analysis}

\begin{table}[t]
\centering
\small
\setlength{\tabcolsep}{3.5pt}
\begin{tabular}{@{}lccccc@{}}
\toprule
\textbf{Model} & \textbf{IFT} & \textbf{0-s} & \textbf{1-s} & \textbf{5-s} & \textbf{10-s} \\
\midrule
\multicolumn{6}{@{}l}{\textbf{Jaccard}} \\
FastAlign            & .144 & \multicolumn{4}{c}{---} \\
ByT5 seq2seq         & .543 & \multicolumn{4}{c}{---} \\
ByT5 pipeline        & .520 & \multicolumn{4}{c}{---} \\
gemma-3n-E4B-it      & .617 & .175 & .257 & .285 & .281 \\
gemma-3-12b-it       & .666 & .238 & .278 & .304 & .314 \\
Qwen3.5-9B           & .693 & \textbf{.296} & \textbf{.377} & \textbf{.376} & \textbf{.378} \\
phi-4                & \textbf{.716} & .092 & .358 & .230 & .299 \\
\addlinespace
\multicolumn{6}{@{}l}{\textbf{Meaning Faithfulness}} \\
FastAlign            & .320 & \multicolumn{4}{c}{---} \\
ByT5 seq2seq         & .550 & \multicolumn{4}{c}{---} \\
ByT5 pipeline        & .552 & \multicolumn{4}{c}{---} \\
gemma-3n-E4B-it      & .746 & .473 & .557 & .572 & .565 \\
gemma-3-12b-it       & .768 & .517 & .557 & .546 & .566 \\
Qwen3.5-9B           & .781 & \textbf{.570} & .631 & \textbf{.628} & \textbf{.636} \\
phi-4                & \textbf{.787} & .182 & \textbf{.635} & .604 & .628 \\
\bottomrule
\end{tabular}
\caption{Glossary generation across tuning and prompting regimes. Jaccard measures
key correctness; Meaning Faithfulness measures semantic accuracy of values against
the ground truth. Both are averaged per entry. $n$-s = $n$-shot prompting.
FastAlign and ByT5-Sanskrit are trained, not prompted, so few-shot settings do not
apply. Leading zeros omitted; best per column in \textbf{bold}.}
\label{tab:f1_1a}
\end{table}
\autoref{tab:f1_1a} reports Jaccard and Meaning Faithfulness across all tuning and prompting regimes. IFT outperforms both \texttt{FastAlign} and in-context learning on every model and metric, with phi-4 best overall. Few-shot prompting improves over zero-shot but plateaus after one example, whereas instruction tuning yields larger and more stable gains --- grounded glossary generation benefits from task-specific adaptation beyond token-level alignment or prompting. Meaning Faithfulness exceeds Jaccard throughout, indicating that models often produce semantically plausible meanings even when segmentation is imperfect.
\autoref{appendix:jaccard-vs-prf1} compares Jaccard against precision, recall,
and F1.
\subsection{Qualitative Analysis}
\begin{table}[t]
\centering
\small
\begin{tabular}{lr}
\toprule
\textbf{Error Category} & \textbf{Count} \\
\midrule
Over-Segmentation & 122 \\
General Segmentation Issues & 22 \\
Both Over and Under Segmentation & 12 \\
Under-Segmentation & 9 \\
Semantic \& Translation Errors & 4 \\
No Issues & 5 \\
\bottomrule
\end{tabular}
\caption{Error analysis findings across 174 generated samples.}
\label{tab:error-analysis}
\end{table}

We perform qualitative error analysis on 174 low-scoring samples (bottom 5\% by Meaning Faithfulness) and categorize errors into six groups (\autoref{tab:error-analysis}). The dominant failure mode is \textit{Over-Segmentation} (70\%), where Sandhi and Sam\={a}sa compounds are incorrectly fragmented, e.g., \texttt{trasare\d{n}u\d{h}}(\devanagari{त्रसरेणुः}) split into \texttt{tra}(\devanagari{त्र}) and \texttt{sare\d{n}u\d{h}}(\devanagari{सरेणुः}).

We also observe concurrent over- and under-segmentation, indicating inconsistent boundary detection, while purely semantic errors are comparatively rare. These results suggest that morphology-aware constraints and improved compound boundary modeling remain critical for Sanskrit glossary generation.
A detailed analysis is available in \autoref{appendix:qual-anal}.

\subsection{Analysis on Seen vs Unseen Gloss Keys}
\autoref{tab:glosskeys-seen-vs-unseen} separates gloss keys seen during
IFT (12{,}194) from those that were not (8{,}956). Gains are
large on both: phi-4 improves by 66 points on seen keys and 60 points on unseen
keys. Memorization of training units would predict little improvement on unseen
keys; the near-parity instead indicates that the models learn generalizable
segmentation and glossing behavior.
\begin{table}[t]
\centering
\small
\setlength{\tabcolsep}{4pt}
\begin{tabular}{@{}lrrrr@{}}
\toprule
 & \multicolumn{2}{c}{\textbf{Seen}} & \multicolumn{2}{c}{\textbf{Unseen}} \\
\cmidrule(lr){2-3} \cmidrule(lr){4-5}
\textbf{Model} & ZS & IFT & ZS & IFT \\
\midrule
gemma-3n-E4B-it & 30.65 & 80.94 & 17.93 & 54.98 \\
gemma-3-12b-it  & 44.94 & 85.40 & 21.79 & 60.29 \\
Qwen3.5-9B      & \textbf{51.36} & 86.30 & \textbf{27.22} & 64.86 \\
phi-4           & 21.53 & \textbf{87.85} & 6.30 & \textbf{66.45} \\
\midrule
ByT5 seq2seq    & --- & 43.33 & --- & 28.24 \\
ByT5 pipeline   & --- & 68.68 & --- & 49.29 \\
\bottomrule
\end{tabular}
\caption{Gloss keys correctly predicted (\%), split by whether the key was seen
during instruction tuning ($N=12{,}194$ seen, $8{,}956$ unseen).
ZS = zero-shot, IFT = instruction fine-tuned. ByT5-Sanskrit models are trained
from scratch and have no zero-shot setting; \emph{pipeline} segments before
glossing. Best per column in \textbf{bold}.}
\label{tab:glosskeys-seen-vs-unseen}
\end{table}
\section{Ablations}

We perform two ablation studies to analyze the role of explicit structural decomposition and robustness to Sanskrit word-order variation on the best performing model, \texttt{phi-4}.

\subsection{Ablation 1: Segmentation-Augmented Glossary Generation}

\begin{table}[t]
\centering
\small
\resizebox{0.482\textwidth}{!}{
\setlength{\tabcolsep}{5pt}
\begin{tabular}{@{}lcc@{}}
\toprule
 & & \textbf{Meaning} \\
\textbf{Model} & \textbf{Jaccard} & \textbf{Faithfulness} \\
\midrule
\multicolumn{3}{@{}l}{\emph{Models that segment and gloss jointly}} \\
gemma-3n-E4B-it & 0.616 & 0.748 \\
gemma-3-12b-it  & 0.656 & 0.769 \\
Qwen3.5-9B      & 0.696 & 0.783 \\
phi-4           & \textbf{0.708} & \textbf{0.785} \\
\addlinespace
\multicolumn{3}{@{}l}{\emph{phi-4 with external segmentation}} \\
\quad + ByT5-Sanskrit (predicted) & 0.501 & 0.741 \\
\quad + gold segments \emph{(oracle)} & 1.000 & \textit{0.872} \\
\bottomrule
\end{tabular}
}
\caption{Performance of segmentation-augmented glossary generation. Models in the
first block segment the \'{s}loka and generate meanings in a single pass; the second
block feeds phi-4 externally produced segmentations. Best system result in
\textbf{bold}; the gold-segment row is an oracle upper bound, not a comparable system.}
\label{tab:ablation1_results}
\end{table}

In this ablation, we instruction fine-tune models to first generate semantic Sanskrit segments and then produce grounded glossary meanings, in order to evaluate the impact of explicit intermediate segmentation. \autoref{tab:ablation1_results} results show that although performance is nearly identical to direct glossary generation for the models selected but it is seen that by providing the gold segmentations to generate only the glossary, we found that the Meaning Faithfulness score increased from 0.787 to 0.872. This indicates that segmentation quality plays an important role not only in recovering the correct key set, but also in improving the semantic faithfulness of the generated glossary. 

\subsection{Ablation 2: Robustness to Word Order Perturbation}
Sanskrit preserves semantic relations through morphology, permitting relatively free word order. To test robustness to this variation, we randomly shuffle words within each \'{s}loka line, keeping the original line boundaries delimited by ``|'' and ``||''.

Comparing \autoref{tab:ablation2_results} against \autoref{tab:f1_1a}, perturbation degrades every model, most severely \texttt{Qwen3.5-9B} (Jaccard $0.693 \rightarrow 0.394$). Even after adaptation, then, models rely on sequential cues rather than the morphological marking.

These results also speak to contamination. The pre-trained models may have seen publicly available Sanskrit texts and analyses, but weak zero-shot performance and the large gains from instruction fine-tuning indicate task-specific adaptation rather than retrieval of memorized material.

\begin{table}[H]
\centering
\small

\begin{tabular}{lcc}
\hline
\textbf{Model} & \textbf{Jaccard} & \makecell{\textbf{Meaning}\\ \textbf{Faithfulness}} \\
\hline
gemma-3-12b-it & 0.587 & 0.750 \\
phi-4        & \textbf{0.632} & \textbf{0.768} \\
Qwen3.5-9B & 0.394 & 0.682 \\   
\hline
\end{tabular}

\caption{Performance under word-order perturbation: words within each \'{s}loka line are randomly shuffled.}
\label{tab:ablation2_results}
\end{table}

\section{Conclusion}
We introduce grounded glossary generation as a structured task requiring recovery of semantically meaningful Sanskrit phrases and translation-grounded glossary meanings. We further present a benchmark dataset, evaluation framework, and systematic study of large language models, showing that instruction fine-tuning substantially improves glossary generation while explicit segmentation supervision affect and word-ordering does affect the performance.

Our results identify morphology and compound boundary detection as the primary bottlenecks for Sanskrit glossary generation. Future work can explore morphology-aware modeling, retrieval-grounded generation, and multilingual lexicographic systems for low-resource settings.


\section*{Limitations}
\begin{itemize}[noitemsep, nolistsep]
\item \textbf{Corpus coverage.} Our data is drawn from two texts, the
V\={a}lm\={\i}ki R\={a}m\={a}ya\d{n}a and \'{S}r\={\i}mad Bh\={a}gavatam.
Both are narrative \textit{itihāsa-purāṇa} works with comparatively regular
syntax; results may not transfer to traditions with distinct compositional
styles, such as ornate K\={a}vya or Vedic prose.

\item \textbf{Metric sensitivity.} Jaccard relies on surface string matching and
Meaning Faithfulness on embedding similarity. Neither recognizes morphological
equivalence between valid sandhi resolutions, so orthographically distinct but
correct segmentations are penalized.

\item \textbf{Training budget.} Instruction tuning ran under memory constraints
with reduced batch sizes and shorter sequence lengths, and all models used LoRA
rather than full fine-tuning. Reported IFT scores are therefore a lower bound on
what the approach can achieve, and whether full fine-tuning closes the remaining
segmentation gap is untested.

\item \textbf{Error analysis scope.} We analyze only the bottom 5\% of samples by
Meaning Faithfulness. Failure modes in the mid-range of the distribution, where
most errors occur, may differ.
\end{itemize}




\section*{Acknowledgments}
We thank Kartikeya Singh (Indian Institute of Technology Kharagpur) for his
contributions to preliminary work on this project. This work was supported in
part by the GCP Research Grant for Gemma and by the National Language
Translation Mission (Bhashini), Government of India. We thank
\texttt{Svarupa}\footnote{\url{https://svarupa.org/}} for providing two
NVIDIA L40 GPUs for this research. We also thank the anonymous reviewers for
feedback that improved the paper.

\bibliography{custom}
\clearpage

\appendix
\section{Appendix}
\label{sec:appendix}
\subsection{Dataset Gloss Key Overlap between Train, Validation and Test}
\begin{table}[H]
    \centering
    \small
    \resizebox{0.48\textwidth}{!}{
    \begin{tabular}{lrrr}
        \toprule
        Split & Unique Gloss Keys & Seen in Train & Percentage \\
        \midrule
        Train       & 101{,}573 & -      & -        \\
        Validation  & 21{,}048  & 12{,}073 & \textbf{57.36\%} \\
        Test        & 21{,}150  & 12{,}194 & \textbf{57.65\%} \\
        \bottomrule
    \end{tabular}
    }
    \caption{Dataset Gloss Key Overlap Information}
    \label{tab:overlap-gloss-key}
\end{table}
\subsection{Hyperparameters used for LLM}
\label{appendix:llm_hyperparameter}

All models were fine-tuned using LoRA with largely shared hyperparameters. Common settings included LoRA scaling ($\alpha=16$), dropout ($0.05$), RS-LoRA, AdamW fused optimization, a learning rate of $3\times10^{-5}$, cosine scheduling with $0.15$ warmup, BF16 precision, gradient clipping ($1.0$), and training for 10 epochs. Checkpointing and evaluation were performed periodically, with the best model selected using validation loss.

Model-specific differences mainly involved LoRA rank, target modules, batch size, and sequence length. \texttt{gemma-3-12b-it} used $r=128$ with adaptation on all linear layers, while \texttt{phi-4} and \texttt{Qwen3.5-9B} used $r=256$. \texttt{Qwen3.5-9B} employed assistant\_only\_loss, whereas the others used completion\_only\_loss. \texttt{gemma-3n-E4B-it} applied LoRA only to attention and feed-forward projection layers, used smaller batch sizes $(2\times4)$ for per-device batch size and gradient accumulation. Due to memory constraints, we disabled gradient checkpointing, and used shorter input sequences (512 tokens) compared to \texttt{gemma-3-12b-it} where we used $(4\times4)$ while \texttt{phi-4} and \texttt{Qwen3.5-9B} used $(8\times2)$.

\subsection{Evaluating selected models Translation capability}
\label{appendix:model-translation-capability}
According to \autoref{tab:model-translation-comparison}, the translation-quality ranking tracks the Jaccard ranking: Phi-4 leads both; Qwen, Gemma-3, and Gemma-3n cluster in the middle. This confirms the hypothesis: \textit{how well the model was exposed to Sanskrit data in its pre-training and to its ability to understand the Sanskrit-to-English translation task shedding further light on the relative strengths of these models}, rather than challenging our contribution i.e. Grounded Glossary Generation performance connects to pre-training Sanskrit exposure, a useful diagnostic property of the benchmark. It does not alter the our core claim that segmentation is the dominant bottleneck, which the ground-truth-substitution experiment isolates independently of translation ability. 
\begin{table}[ht]
\centering
\small
\resizebox{0.48\textwidth}{!}{
\begin{tabular}{lccc}
\hline
\textbf{Model} & \textbf{BLEU} & \textbf{ChrF} & \textbf{Semantic Similarity} \\
\hline
Phi-4          & 6.21 & 30.61 & 73.77\% \\
Qwen3.5-9B     & 4.96 & 28.11 & 70.20\% \\
Gemma-3n-E4B-it & 4.93 & 27.70 & 70.57\% \\
Gemma-3-12B-it & 4.69 & 27.31 & 70.83\% \\
\hline
\end{tabular}
}
\caption{Comparison of model performance across BLEU, ChrF, and semantic similarity.}
\label{tab:model-translation-comparison}
\end{table}

\subsection{Intuition behind the design of Meaning Faithfulness} 
\label{appendix:metric-intuition}
The intuition behind this design of metric is to explicitly handle three common scenarios in key generation: under segmentation, over segmentation and proper segmentation.

\noindent \textbf{Under Segmentation:}
In this case, the model fails to split a word that should have been fragmented. Since the ground-truth key is segmented, the maximum Levenshtein similarity between the generated key and the ground truth keys typically decreases. Multiplying this lower KeySim with the cosine similarity therefore penalizes the generated key.

\noindent \textbf{Over segmentation:}
Here, the model splits a key into fragments more than necessary. As in the previous case, the resulting fragments obtain lower Levenshtein similarity scores with the ground-truth keys, and multiplying by KeySim reduces their contribution, thereby penalizing the model.

\noindent In both of the above cases, severe under segmentation or over segmentation may cause the similarity score to fall below the threshold, setting KeySim to 0. (For example, this would happen if the model splits a single word into three fragments or merges three separate words into one.)

\noindent \textbf{Proper segmentation:} When the key is segmented correctly, KeySim remains high, 100\% for an exact match and close to 100\% for near perfect matches. In such cases, the key is effectively not penalized.


\subsection{Qualitative Analysis of Phi-4}
\label{appendix:qual-anal}
To better understand model failures beyond quantitative metrics, we conducted a qualitative error analysis on 174 representative samples (bottom 5\% by Meaning Faithfulness score). We grouped the errors into six mutually exclusive categories based on mismatches between the reference and generated glossary entries (\autoref{tab:error-analysis}).

The most frequent issue was \textit{Over-Segmentation} (70\% of cases), where compounds formed through Sandhi and Samasa were fragmented into smaller units, leading to loss of contextual meaning. For example, \texttt{trasare\d{n}u\d{h}}(\devanagari{त्रसरेणुः}) was incorrectly split into \texttt{tra}(\devanagari{त्र}) and \texttt{sare\d{n}u\d{h}}(\devanagari{सरेणुः}).

We also observed cases of \textit{Concurrent Over and Under Segmentation}, where the model simultaneously failed to split some compounds while over-splitting adjacent phrases, indicating inconsistent boundary detection in longer Sanskrit strings. \textit{Under-Segmentation} was less common but still harmful, as in \texttt{puru\d{s}a-paricaryay={a}}(\devanagari{पुरुष-परिचर्यया}), which remained unsplit despite the reference expecting finer segmentation.

A smaller subset of errors were purely \textit{Semantic}, where the generated gloss was contextually incorrect, such as translating \texttt{amba\d{s}\d{t}ha}(\devanagari{अम्बष्ठ}) as `driver'' instead of `elephant-keeper.'' 

For the 5 no issues samples, there were no issues on the generated output but rather the key:val pairs were wrongly translated in the ground truth glossary.  

Overall, these findings highlight the need for stronger morphological constraints and improved boundary detection to reduce over-fragmentation in Sanskrit glossary generation.

\subsection{Comparing Jaccard against Precision, Recall, F1-score}
\label{appendix:jaccard-vs-prf1}
Precision, Recall, are also valid alternatives for evaluating glossary-unit recovery. We chose Jaccard because the task is formulated as recovering the set of glossary units, and Jaccard provides a simple set-overlap measure that jointly penalizes missing and spurious units. Since Jaccard and F1 are closely related monotonic measures, we do not expect the relative ranking of models to change substantially which is also observed in \autoref{tab:jaccard-vs-prf1}. Similarly for Ablation 1, the Jaccard vs Precision, Recall and F1-score, the respective scores are available in \autoref{tab:jaccard-vs-prf1-abl1}.
\begin{table}[h]
    \centering
    \small
    \resizebox{0.48\textwidth}{!}{
    \begin{tabular}{lcccc}
        \toprule
        \textbf{Model} & \textbf{Precision} & \textbf{Recall} & \textbf{F1-score} & \textbf{Jaccard} \\
        \midrule
        Phi-4         & 0.8145 & 0.8237 & 0.8175 & 0.716 \\
        Qwen3.5-9B    & 0.8016 & 0.8041 & 0.8012 & 0.693 \\
        Gemma-3-12B-it & 0.7775 & 0.7875 & 0.7807 & 0.666 \\
        Gemma-3n-E4B-it & 0.7439 & 0.7453 & 0.7426 & 0.617 \\
        \bottomrule
    \end{tabular}
    }
    \caption{The metric values for our test data, for correct gloss key identification}
    \label{tab:jaccard-vs-prf1}
\end{table}

\begin{table}[h]
    \centering
    \small
    \resizebox{0.48\textwidth}{!}{
    \begin{tabular}{lcccc}
        \toprule
        \textbf{Model} & \textbf{Precision} & \textbf{Recall} & \textbf{F1-score} & \textbf{Jaccard} \\
        \midrule
        Phi-4 & 0.8102 & 0.8173 & 0.8120 & 0.708 \\
        Qwen3.5-9B & 0.8055 & 0.8042 & 0.8031 & 0.696 \\
        Gemma-3-12B-it & 0.7770 & 0.7732 & 0.7732 & 0.656 \\
        Gemma-3n-E4B-it & 0.745 & 0.742 & 0.743 & 0.616 \\
        \bottomrule
    \end{tabular}
    }
    \caption{Ablation 1 metric values for our test data, for correct gloss key identification}
    \label{tab:jaccard-vs-prf1-abl1}
\end{table}
\subsection{Promtps used for N-shot and IFT}
\begin{tcolorbox}[title=System Prompt, colback=white, colframe=orange]
\small
\label{box:system_prompt}
    You are a Sanskrit-to-English glossary extraction engine. Given a Sanskrit shloka and its English translation, output a single JSON object mapping each Sanskrit word to its English meaning.

Rules: \\
1. Output ONLY a Python dict. Start with "\{", end with "\}". No other text. \\
2. Keys are Sanskrit words (or sandhi-resolved tokens) as they appear in the shloka. \\
3. Values are their English meanings as evidenced by the translation. \\
4. No markdown, no code fences, no explanation, no preamble. \\

Format: \{"glossary": \{"sanskrit\_word": "english\_meaning", ...\}\}
\end{tcolorbox}

\begin{tcolorbox}[title=User Prompt, colback=white, colframe=red]
\small
\label{box:user_prompt}
Generate a Sanskrit-English glossary mapping for the following shloka based on its translation.

Shloka:
\{shloka\}

Translation:
\{translation\}
\end{tcolorbox}
\subsection{Prompts used for Ablation 1}
\label{appendix:prompt-ablation-1}
\begin{tcolorbox}[title=System Prompt for Ablation 1, colback=white, colframe=green!35!brown]
\small
    Perform Padaccheda (resolve sandhi, keep samasa compound words intact). Analyze the shloka word by word using the translation as a reference. End your reasoning with the resolved word sequence as: [word1, word2, ...]. \\

\#\# STEP 2 — FINAL GENERATION: \\
Output the glossary as a code block. Rules for the output: \\
1. Output exactly ONE valid Python dict with a single key 'glossary'. \\
2. All keys are the Padaccheda-resolved Sanskrit words, in the order they appear in the shloka. \\
3. All values are their English meanings, derived strictly from the provided translation. \\
4. No trailing commas. \\
5. No extra keys, no nested objects — flat dictionary only. \\

\#\# RESPONSE FORMAT (follow exactly, do not deviate): \\
\#\#\# REASONING: \\
<your padaccheda analysis ending with the resolved word list> \\

\#\#\# FINAL GENERATION: \\
\{``glossary'': \{``sanskrit\_word'': ``english\_meaning'', ...\}\}\\
Do NOT include any text outside these two sections. Do NOT add greetings, notes, or explanations after the output block.
\end{tcolorbox}
\begin{tcolorbox}[title=User Prompt for Ablation 1 IFT, colback=white, colframe=green!35!brown]
\small
   \#\#\# INPUT:
You are an expert in Sanskrit. Perform Padaccheda (word resolution) on the following Sanskrit shloka, then generate the glossary mapping using the generated Padaccheda and provided English translation.

Shloka:
<shloka>

Translation:
<translation>

\end{tcolorbox}

\subsection{N-shot Examples}
\label{appendix:n-shot-example}
The following data was used as examples for all N-shot experiments. For 1-shot, the first example, for 5-shot, the first 5 examples and similarly for 10-shot all the 10 examples were used.
\begin{tcolorbox}[breakable]
\small
Shloka: \devanagari{न वयं क्लेशबीजानि यत: स्यु: पुरुषर्षभ ।
पुरुषं तं विजानीमो वाक्यभेदविमोहिता: । }\\
Translation: O greatest among human beings, it is very difficult to ascertain the particular miscreant who has caused our sufferings, because we are bewildered by all the different opinions of theoretical philosophers. \\
Glossary: \{\devanagari{'न'}: 'not', \devanagari{'वयम्'}: 'we', \devanagari{'क्लेश-बीजानि'}: 'the root cause of sufferings', \devanagari{'यतः'}: 'wherefrom', \devanagari{'स्युः'}: 'it so happens', \devanagari{'पुरुष-ऋषभ'}: 'O greatest of all human beings', \devanagari{'पुरुषम्'}: 'the person', \devanagari{'तम्'}: 'that', \devanagari{'विजानीमः'}: 'know', \devanagari{'वाक्य-भेद'}: 'difference of opinion', \devanagari{'विमोहिताः'}: 'bewildered by'\} \\

Shloka: \devanagari{सोऽधिक्षिप्तो दुर्वाचोभि: पदाहत इवोरग: ।
निश्चक्राम गदापाणिरमर्षात्ताम्रलोचन: ।} \\
Translation: Offended by these harsh words, \'{s}ambara became as agitated as a kicked snake. He came out, club in hand, his eyes red with rage. \\
Glossary: \{\devanagari{'सः'}: 'he', \devanagari{'अधिक्षिप्तः'}: 'insulted', \devanagari{'दुर्वाचोभिः'}: 'by harsh words', \devanagari{'पदा'}: 'by a foot', \devanagari{'आहतः'}: 'struck', \devanagari{'इव'}: 'like', \devanagari{'उरगः'}: 'a snake', \devanagari{'निश्चक्राम'}: 'came out', \devanagari{'गदा'}: 'a club', \devanagari{'पाणिः'}: 'in his hand', \devanagari{'अमर्षात्'}: 'out of intolerant anger', \devanagari{'ताम्र'}: 'copper-red', \devanagari{'लोचनः'}: 'whose eyes'\} \\

Shloka:\devanagari{ दृष्ट्वान्यांश्च महोत्पातानतत्तत्त्वविद: प्रजा: ।
ब्रह्मपुत्रानृते भीता मेनिरे विश्‍वसम्प्लवम् ।} \\
Translation: Marking these and many other omens of evil times, everyone but the four sage sons of Brahm\={a}, who were aware of the fall of Jaya and Vijaya and of their birth as Diti's sons, was seized with fear. They did not know the secrets of these portents and thought that the dissolution of the universe was at hand. \\
Glossary: \{\devanagari{'दृष्ट्वा'}: 'having seen', \devanagari{'अन्यान्'}: 'others', \devanagari{'च'}: 'and', \devanagari{'महा'}: 'great', \devanagari{'उत्पातान्'}: 'evil omens', \devanagari{'अ-तत्-तत्त्व-विदः'}: 'not knowing the secret (of the portents)', \devanagari{'प्रजाः'}: 'people', \devanagari{'ब्रह्म-पुत्रान्'}: 'the sons of Brahm\={a} (the four Kum\={a}ras)', \devanagari{'ऋते'}: 'except', \devanagari{'भीताः'}: 'being fearful', \devanagari{'मेनिरे'}: 'thought', \devanagari{'विश्व-सम्प्लवम्'}: 'the dissolution of the universe'\} \\

Shloka: \devanagari{कथं वर्तेत विहरेत् कैर्वा ज्ञायेत लक्षणै: ।
किं भुञ्जीतोत विसृजेच्छयीतासीत याति वा ।
एतदच्युत मे ब्रूहि प्रश्न‍ं प्रश्न‍‌विदां वर ।
नित्यबद्धो नित्यमुक्त एक एवेति मे भ्रम: ।} \\
Translation: O my Lord, Acyuta, the same living entity is sometimes described as eternally conditioned and at other times as eternally liberated. I am not able to understand, therefore, the actual situation of the living entity. You, my Lord, are the best of those who are expert in answering philosophical questions. Please explain to me the symptoms by which one can tell the difference between a living entity who is eternally liberated and one who is eternally conditioned. In what various ways would they remain situated, enjoy life, eat, evacuate, lie down, sit or move about? \\
Glossary: \{\devanagari{'कथम्'}: 'in what way', \devanagari{'वर्तेत'}: 'he is situated', \devanagari{'विहरेत्'}: 'he enjoys', \devanagari{'कैः'}: 'by which', \devanagari{'वा'}: 'or', \devanagari{'ज्ञायेत'}: 'would be known', \devanagari{'लक्षणैः}': 'by symptoms', \devanagari{'किम्'}: 'what', \devanagari{'भुञ्जीत'}: 'he would eat', \devanagari{'उत'}: 'and', \devanagari{'विसृजेत्'}: 'would evacuate', \devanagari{'शयीत'}: 'would lie down', \devanagari{'आसीत'}: 'would sit', \devanagari{'याति'}: 'goes', \devanagari{'एतत्'}: 'this', \devanagari{'अच्युत'}: 'O Acyuta', \devanagari{'मे'}: 'to me',\devanagari{'ब्रूहि'}: 'explain', \devanagari{'प्रश्नम्'}: 'the question', \devanagari{'प्रश्न-विदाम्'}: 'of all those who know how to answer questions', \devanagari{'वर'}: 'O the best', \devanagari{'नित्य-बद्धः'}: 'eternally conditioned', \devanagari{'नित्य-मुक्तः'}: 'eternally liberated', \devanagari{'एकः'}: 'singular', \devanagari{'एव'}: 'certainly', \devanagari{'इति'}: 'thus', \devanagari{'भ्रमः'}: 'confusion'\} \\

Shloka: \devanagari{तत्रेतिकृत्यमुपशिक्ष यथोपदेशं
येनैष मे कर्शितोऽतिरिरंसयात्मा ।
सिद्ध्येत ते कृतमनोभवधर्षिताया
दीनस्तदीश भवनं सद‍ृशं विचक्ष्व ।} \\
Translation: Devah\={u}ti continued: My dear lord, I am struck by excited emotion for you. Therefore kindly make what arrangements must be made according to the scriptures so that my skinny body, emaciated through unsatisfied passion, may be rendered fit for you. Also, my lord, please think of a suitable house for this purpose. \\
Glossary: \{\devanagari{'तत्र'}: 'in that', \devanagari{'इति-कृत्यम्'}: 'what is necessary to be done', \devanagari{'उपशिक्ष'}: 'perform', \devanagari{'यथा'}: 'according to', \devanagari{'उपदेशम्'}: 'instruction in scripture', \devanagari{'येन'}: 'by which', \devanagari{'एषः'}: 'this', \devanagari{'मे'}: 'my', \devanagari{'कर्शितः'}: 'emaciated', \devanagari{'अतिरिरꣳ-सया'}: 'due to intense passion not being satisfied', \devanagari{'आत्मा'}: 'body', \devanagari{'सिद्ध्येत'}: 'it may be rendered fit', \devanagari{'ते'}: 'for you', \devanagari{'कृत'}: 'excited', \devanagari{'मनः-भव'}: 'by emotion', \devanagari{'धर्षितायाः'}: 'who am struck', \devanagari{'दीनः'}: 'poor', \devanagari{'तत्'}: 'therefore', \devanagari{'ईश'}: 'O my dear lord', \devanagari{'भवनम्'}: 'house', \devanagari{'सदृशम्'}: 'suitable', \devanagari{'विचक्ष्व'}: 'please think of'\}

Shloka: \devanagari{न वेद धर्मं किल पद्मयोनि-
र्न ब्रह्मपुत्रा भृगुनारदाद्या: ।
न वै कुमार: कपिलो मनुश्च
ये नो निषेधन्त्यतिवर्तिनं हरम् ।} \\
Translation: Alas, Lord Brahm\={a}, who has taken his birth from the lotus flower, does not know the principles of religion, nor do the great saints like Bhṛgu and N\={a}rada, nor the four Kum\={a}ras, headed by Sanat-kum\={a}ra. Manu and Kapila have also forgotten the religious principles. I suppose it to be because of this that they have not tried to stop Lord \'{s}iva from behaving improperly. \\
Glossary: \{ \devanagari{'न'}: 'not', \devanagari{'वेद'}: 'knows', \devanagari{'धर्मम्'}: 'the religious principles', \devanagari{'किल'}: 'indeed', \devanagari{'पद्म-योनिः'}: 'Lord Brahm\={a}', \devanagari{'ब्रह्म-पुत्राः'}: 'the sons of Lord Brahm\={a}', \devanagari{'भृगु'}: 'Bhṛgu', \devanagari{'नारद'}: 'N\={a}rada', \devanagari{'आद्याः'}: 'and so on', \devanagari{'वै'}: 'indeed', \devanagari{'कुमारः'}: 'the four Kum\={a}ras (Sanaka', \devanagari{'कपिलः'}: 'Lord Kapila', \devanagari{'मनुः'}: 'Manu himself', \devanagari{'च'}: 'and', \devanagari{'ये'}: 'who', \devanagari{'नो'}: 'not', \devanagari{'निषेधन्ति'}: 'order to stop', \devanagari{'अति-वर्तिनम्'}: 'who is beyond laws and orders', \devanagari{'हरम्'}: 'Lord \'{s}iva'\}\\

Shloka:\devanagari{ कर्माण्यारभते येन पुमानिह विहाय तम् ।
अमुत्रान्येन देहेन जुष्टानि स यदश्नुते ।} \\
Translation: The results of whatever a living entity does in this life are enjoyed in the next life. \\
Glossary: \{\devanagari{'कर्माणि'}: 'fruitive activities', \devanagari{'आरभते'}: 'begins to perform', \devanagari{'येन'}: 'by which', \devanagari{'पुमान्'}: 'a living entity', \devanagari{'इह'}: 'in this life', \devanagari{'विहाय'}: 'giving up', \devanagari{'तम्'}: 'that', \devanagari{'अमुत्र'}: 'in the next life', \devanagari{'अन्येन'}: 'another', \devanagari{'देहेन'}: 'by a body', \devanagari{'जुष्टानि'}: 'the results', \devanagari{'सः'}: 'he', \devanagari{'यत्'}: 'that', \devanagari{'अश्नुते'}: 'enjoys'\} \\

Shloka: \devanagari{आरोप्यारुरुहे यानं भ्रातृभ्यां हनुमद्युत: ।
विभीषणाय भगवान् दत्त्वा रक्षोगणेशताम् ।
लङ्कामायुश्च कल्पान्तं ययौ चीर्णव्रत: पुरीम् । }\\
Translation: After giving Vibh\={i}ṣaṇa the power to rule the R\={a}kṣasa population of La\.{n}k\={a} for the duration of one kalpa, Lord R\={a}macandra, the Supreme Personality of Godhead [Bhagav\={a}n], placed S\={i}t\={a}dev\={i} on an airplane decorated with flowers and then got on the plane Himself. The period for His living in the forest having ended, the Lord returned to Ayodhy\={a}, accompanied by Hanum\={a}n, Sugr\={i}va and His brother Lakṣmaṇa. \\
Glossary: \{\devanagari{'आरोप्य'}: 'keeping or placing', \devanagari{'आरुरुहे'}: 'got up', \devanagari{'यानम्'}: 'on the airplane', \devanagari{'भ्रातृभ्याम्'}: 'with His brother Lakṣmaṇa and the commander Sugr\={i}va', \devanagari{'हनुमत्-युतः'}: 'accompanied by Hanum\={a}n', \devanagari{'विभीषणाय'}: 'unto Vibh\={i}ṣaṇa', \devanagari{'भगवान्'}: 'the Lord', \devanagari{'दत्त्वा'}: 'gave charge', \devanagari{'रक्षः-गण-ईशताम्'}: 'the power to rule over the R\={a}kṣasa population of La\.{n}k\={a}', \devanagari{'लङ्काम्'}: 'the state of La\.{n}k\={a}', \devanagari{'आयुः च'}: 'and the duration of life', \devanagari{'कल्प-अन्तम्'}: 'for many', \devanagari{'ययौ'}: 'returned home', \devanagari{'चीर्ण-व्रतः'}: 'finishing the duration of time living in the forest', \devanagari{'पुरीम्'}: 'to Ayodhy\={a}-pur\={i}'\} \\

Shloka: \devanagari{एतां विद्यामधिगतो विश्वरूपाच्छतक्रतु: ।
त्रैलोक्यलक्ष्मीं बुभुजे विनिर्जित्य मृधेऽसुरान् । }\\
Translation: King Indra, who performed one hundred sacrifices, received this prayer of protection from Vi\'{s}var\={u}pa. After conquering the demons, he enjoyed all the opulences of the three worlds. \\
Glossary: \{\devanagari{'एताम्'}: 'this', \devanagari{'विद्याम्'}: 'prayer', \devanagari{'अधिगतः'}: 'received', \devanagari{'विश्वरूपात्'}: 'from the br\={a}hmaṇa Vi\'{s}var\={u}pa', \devanagari{'शत-क्रतुः'}: 'Indra', \devanagari{'त्रैलोक्य-लक्ष्मीम्'}: 'all the opulence of the three worlds', \devanagari{'बुभुजे'}: 'enjoyed', \devanagari{'विनिर्जित्य'}: 'conquering', \devanagari{'मृधे'}: 'in battle', \devanagari{'असुरान्'}: 'all the demons'\} \\

Shloka: \devanagari{ननृतुस्तस्य पुरत: स्त्रियोऽथो गायका जगु: ।
मृदङ्गवीणापणवैर्वाद्यं चक्रुर्मनोरमम् ।} \\
Translation: The women danced before the sage, and the celestial singers sang to the charming accompaniment of drums, cymbals and v\={i}ṇ\={a}s. \\
Glossary: \{\devanagari{'ननृतुः'}: 'danced', \devanagari{'तस्य'}: 'of him', \devanagari{'पुरतः'}: 'in front', \devanagari{'स्त्रियः'}: 'women', \devanagari{'अथ उ'}: 'and furthermore', \devanagari{'गायकाः'}: 'singers', \devanagari{'जगुः'}: 'sang', \devanagari{'मृदङ्ग'}: 'with drums', \devanagari{'वीणा'}: 'stringed instruments', \devanagari{'पणवैः'}: 'and cymbals', \devanagari{'वाद्यम्'}: 'instrumental music', \devanagari{'चक्रुः'}: 'they made', \devanagari{'मनः-रमम्'}: 'charming'\}
\end{tcolorbox}

\end{document}